\documentclass[10pt,twocolumn,letterpaper]{article}
\usepackage[accsupp]{axessibility} 

\usepackage{wacv}
\usepackage{times}
\usepackage{epsfig}
\usepackage{graphicx}
\usepackage{amsmath}
\usepackage{amssymb}
\usepackage{booktabs}
\usepackage{color}
\usepackage{comment}
\usepackage[usenames,dvipsnames]{xcolor}
\usepackage{algorithm}
\usepackage{algorithmic}
\usepackage{authblk}

\usepackage[cjk]{kotex} 

\usepackage{multirow}
\usepackage{comment}
\usepackage{newfloat}
\usepackage{listings}
\usepackage{mathtools}
\usepackage{xspace}

\newcommand{\mbc}{\mathbf{c}}

\newcommand{\mbe}{\mathbf{e}}

\newcommand{\mbh}{\mathbf{h}}

\newcommand{\mbt}{\mathbf{t}}

\newcommand{\mbx}{\mathbf{x}}

\newcommand{\ignore}[1]{}

\makeatletter
\DeclareRobustCommand\onedot{\futurelet\@let@token\@onedot}
\def\@onedot{\ifx\@let@token.\else.\null\fi\xspace}

\def\etal{{et al}\onedot}
\makeatother

\newcommand\blfootnote[1]{%
  \begingroup
  \renewcommand\thefootnote{}\footnote{#1}%
  \addtocounter{footnote}{-1}%
  \endgroup
}

\def\wacvPaperID{1444} 

\wacvalgorithmstrack   

\wacvfinalcopy 

\ifwacvfinal
\usepackage[breaklinks=true,bookmarks=false]{hyperref}
\else
\usepackage[pagebackref=true,breaklinks=true,colorlinks,bookmarks=false]{hyperref}
\fi

\begin{document}

\title{Image-free Domain Generalization via CLIP for 3D Hand Pose Estimation}
\author{Seongyeong~Lee\textsuperscript{1,2\dag} \qquad 
Hansoo~Park\textsuperscript{1} \qquad 
Dong~Uk~Kim\textsuperscript{1} \qquad 
Jihyeon~Kim\textsuperscript{1} \qquad 
Muhammadjon~Boboev\textsuperscript{1} \qquad
Seungryul~Baek\textsuperscript{1} \qquad
}
\affil{\textsuperscript{1}UNIST, South Korea \qquad \textsuperscript{2}NC Soft, South Korea}

\maketitle
\blfootnote{This research was conducted when Seongyeong Lee was a graduate student (Master candidate) at UNIST$\dag$.}

\thispagestyle{empty}

\begin{abstract}
RGB-based 3D hand pose estimation has been successful for decades thanks to large-scale databases and deep learning. However, the hand pose estimation network does not operate well for hand pose images whose characteristics are far different from the training data. This is caused by various factors such as illuminations, camera angles, diverse backgrounds in the input images, etc. Many existing methods tried to solve it by supplying additional large-scale unconstrained/target domain images to augment data space; however collecting such large-scale images takes a lot of labors. In this paper, we present a simple image-free domain generalization approach for the hand pose estimation framework that uses only source domain data. We try to manipulate the image features of the hand pose estimation network by adding the features from text descriptions using the CLIP (Contrastive Language-Image Pre-training) model. The manipulated image features are then exploited to train the hand pose estimation network via the contrastive learning framework. In experiments with STB and RHD datasets, our algorithm shows improved performance over the state-of-the-art domain generalization approaches. 
\end{abstract}

\section{Introduction}
\label{sec:intro}

3D hand pose estimation has been essential for human-machine interaction applications such as augmented and virtual reality and robotics. Depth-based 3D hand pose estimation~\cite{baek2018augmented,ren2022mining,chen2020pose,ge2018hand,huang2020awr,moon2018v2v,wan2018dense,xiong2019a2j,chen2019so,ge2018point,ren2021spatial,cheng2022efficient,cheng2021handfoldingnet,wan2020dual,fang2020jgr} has been popular using the Kinect sensor. Recently, RGB-based 3D hand pose estimation has been widely used and developed thanks to its simple and practical setup compared to depth-based approaches. Many researchers have tried to improve the performance of RGB-based hand pose estimation either by developing the novel deep learning architectures~\cite{zimmermann2019FreiHAND, zimmerman_iccv2017, hampali2020honnotate} or by increasing data via additional data collection or real and synthetic data generation. For example, Lin~\etal~\cite{lin2021mesh} recently proposed a Transformer-based model and showed improvement in hand pose estimation performance. Although the model shows the excellent performance, it is still challenging to secure the generalization ability if there is a huge gap in perspective or severe occlusions~\cite{anil2020generalization}. Using additional data apparently helps improve the generalization ability~\cite{cha2021self2d, binod2020sampling}. However, it is challenging to collect RGB pose data in diverse environments with varied poses securing the accurate annotations, and thus the performance is limited for in-the-wild environments. 

\begin{figure}[!t]
\centering
\includegraphics[width=\linewidth]{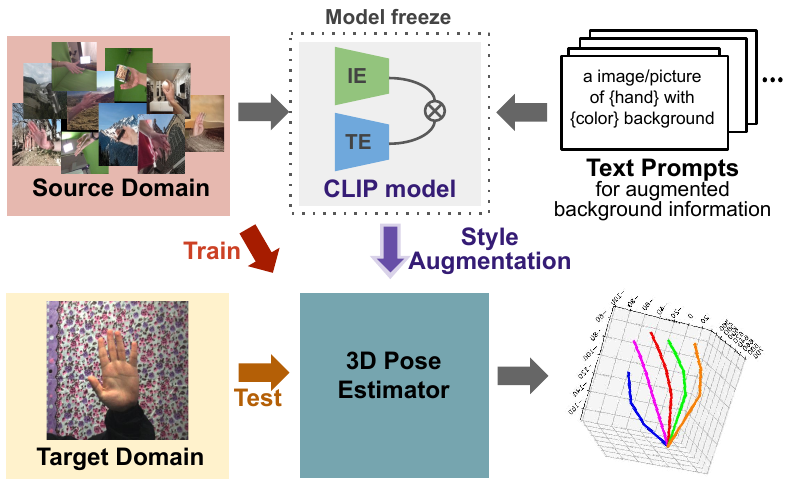}
\caption{Overview of our domain generalization method in the hand pose estimation task exploiting the CLIP model. Given a source domain image, we train the 3D hand pose estimator by exploiting the CLIP text manipulated features as the new features to the pose estimation network.}
\label{fig:myfigure1}
\end{figure}

Due to the challenges in scaling up the dataset size, there have appeared research trends that propose the algorithm for efficient data usage. As an example, various domain adaptation and generalization techniques~\cite{ganin2016domain, hoffman2018cycada, long2018conditional, saito2017adversarial, saito2019semi, jiang2020bidirectional, kim2020attract,yang2020fda, xu2019self, zhang2019category, zou2018unsupervised,NEURIPS2019_2974788b, seo2020learning, yue2019domain} have been proposed. Also, studies that implements the self-supervision via the contrastive learning~\cite{chen2020simple, spurr2021peclr} is also shown in the literature. For generalization beyond the training domain, Zhang~\etal~\cite{zhang2021learning} proposed a method for causal representation learning which explicitly exploits the causal structure of the task. By applying domain adaptation and generalization techniques, they achieved better generalization ability on hand pose estimation domain. In addition, Spurr~\etal~\cite{spurr2021peclr} presented a methodology that enables self-supervised contrastive learning for hand pose estimation with many unlabeled data. They encouraged equal variances for geometric transformations during representation learning. As a result, they improved generalization between datasets using the above methodology.

In this paper, we aim to improve domain generalization performance by efficiently using the features of a given data  rather than using additional datasets. Especially, we proposed to use the CLIP~\cite{{radford2021learning}} model to extract generalized feature representation from the source domain images. The CLIP model used a large number of image-text pairs and contrastive learning for training. So, it can extract a more generalized feature representation than models using only images. Using this advantage, we improve the generalization ability across unseen domains by providing the text encoder of CLIP with various texts for augmentation. To the best of our knowledge, this is the first work to adapt domain generalization in hand pose estimation domain using only the source dataset. An overview of our pose estimator with CLIP is shown in Figure~\ref{fig:myfigure1}.


\section{Related work}

In this section, we will review recent literature on hand pose estimation, contrastive learning and domain adaptation/generalization.

\subsection{3D hand pose estimation}
Hand pose estimation is the task of estimating the x, y and z coordinates of $21$ hand joints from either depth maps~\cite{garcia2018first, malik2020handvoxnet} or RGB images~\cite{baek2019mesh, hasson2019learning}. The depth map is invariant to the lighting conditions and shadows. It also has the advantage of being strong against clutter~\cite{baek2018augmented, malik2020handvoxnet}; while it has the disadvantage of not being able to capture various features such as textures and colors of the scene. The RGB image has the advantage of being able to capture detailed hand attributes such as unique colors, textures, and outlines. Besides, RGB cameras are more ubiquitous than depth sensors in our daily life, as we are holding RGB cameras in our smart phones; while we are not holding depth sensors in our pocket. However, unlike the depth map, the RGB image completely loses the 3D depth information. Therefore, compared to the depth map, there is much difficulty in achieving 3D hand pose estimation from RGB images~\cite{baek2020weakly, hasson2019learning}. The useful yet challenging setting of RGB-based 3D hand pose estimation recently accelerates the development of many algorithms~\cite{moon2020interhand2, lin2021two, zimmermann2017learning, dukim2021twohand}. More recently, RGB-based 3D hand mesh estimation has been established as well~\cite{hasson2019learning, baek2019mesh}. The 3D hand model (ie. MANO)~\cite{baek2019mesh, baek2020weakly, kulon2020weakly} has been exploited when constructing these pipelines. The graph convolutional network (GCN) is further exploited to better capture the relationship between vertices in the mesh topology~\cite{zhou2020monocular, doosti2020hope}.

In the aspect of data, hands require data collection with much diverse camera perspectives, poses and shapes compared to other domains such as body pose estimation~\cite{cha2022actiontransformer, cha2022multiperson, baek2017kinematicrf, baek2017onlinerf, saqlain20223dmeshgar}. This is due to the fact that hands exhibit severe self-occlusions and diverse camera perspectives.

\subsection{Contrastive learning}

Contrastive learning is a way to achieving the self-supervised learning. It has been utilized in various works~\cite{henaff2020data, oord2018representation, chen2020simple, he2020momentum, tian2020contrastive} for extracting the view-invariant representation from multiple views by exploiting the different views of the same content as positive samples and other content as negative samples. Chen~\etal~\cite{chen2020simple} proposed the contrastive learning for the data augmentation to learn better representation. They used data-augmented samples as positive; while other data samples as negative. The performance of these approaches could be improved by increasing the number of negative samples. However, the number of negative samples is limited by the GPU memory size. So, \cite{chen2020improved, he2020momentum} proposed a method to increase the number of negative samples with the momentum encoder. Zhu~\etal~\cite{zhu2020eqco} proposed a method for controlling the margin term according to the number of negative samples. Caron~\etal~\cite{caron2020unsupervised} addressed the limitation by clustering the augmented data instead of comparing features. Contrastive learning has been also used in various tasks such as image and video classification~\cite{chen2020simple, he2020momentum, tian2020contrastive} and object detection~\cite{henaff2020data}. 

Recently, Spurr~\etal~\cite{spurr2021peclr} proposed the contrastive learning framework for the 3D hand pose estimation task. They extended the SimCLR~\cite{chen2020simple} framework to be applicable to structured regression tasks such as the hand pose estimation. They used the geometric and appearance transformed hand images as positive and other images as negative samples for achieving the contrastive learning. 

Most contrastive learning methods have been dependent on the data augmentation techniques such as scale manipulation, cut out, noise and rotation transformation. In our paper, we use the CLIP model, which is robust to the zero-shot learning and often used for the domain generalization tasks. We augment the data space in terms of style by manipulating the text prompts in CLIP and perform the contrastive learning between features of original and augmented data to make the feature of original data robust to domain generalization.

\subsection{Domain adaptation/generalization}

One of the important problems in computer vision is the dataset bias and domain shift between different datasets. To resolve this problem, domain adaptation (DA) /domain generalization (DG) approaches have recently attracted a lot of attention.

Domain adaptation (DA) is a methodology used in situations where the label of the target domain does not exist or is insufficient. DA utilizes source domain data, sparsely labeled, unlabeled target domain data as the training dataset to effectively learn the target domain distribution. DA is largely divided into three types: supervised domain adaptation (SDA)~\cite{ganin2016domain, hoffman2018cycada, long2018conditional, saito2017adversarial}, semi-supervised domain adaptation (SSDA)~\cite{saito2019semi, jiang2020bidirectional, kim2020attract} and unsupervised domain adaptation (UDA)~\cite{yang2020fda, xu2019self, zhang2019category, zou2018unsupervised}. SDA is a method that uses both labeled source and target domain data, SSDA is a method that exploits source domain data whose label is completely annotated and target domain data whose label is partially annotated. Finally, UDA is a method that uses source domain and target domain data without any labels. 

Above mentioned methods (ie. SDA, SSDA and UDA) all require the target domain data to train the model. However, it is non-trivial to obtain the target data and also collecting the target data with labels is even harder. Domain generalization (DG)~\cite{NEURIPS2019_2974788b, seo2020learning, yue2019domain, zhang2021learning} has emerged to relieve the challenge from collecting target data. DG differs from DA in that it makes features generalized in the target domain by using the source and additional unseen domain data which might be similar to the target domain data. Recently, Zhang~\etal~\cite{zhang2021learning} proposed the DG methodology of the pose estimation using unconstrained dataset and could improve the generalization ability. However, it is limited by the fact that it requires to collect additional unseen domain data which might be similar to the target domain data. In our paper, we propose the domain generalization method which does not require additional dataset collection.

\begin{figure*}[ht]
\centering
\includegraphics[width=\linewidth]{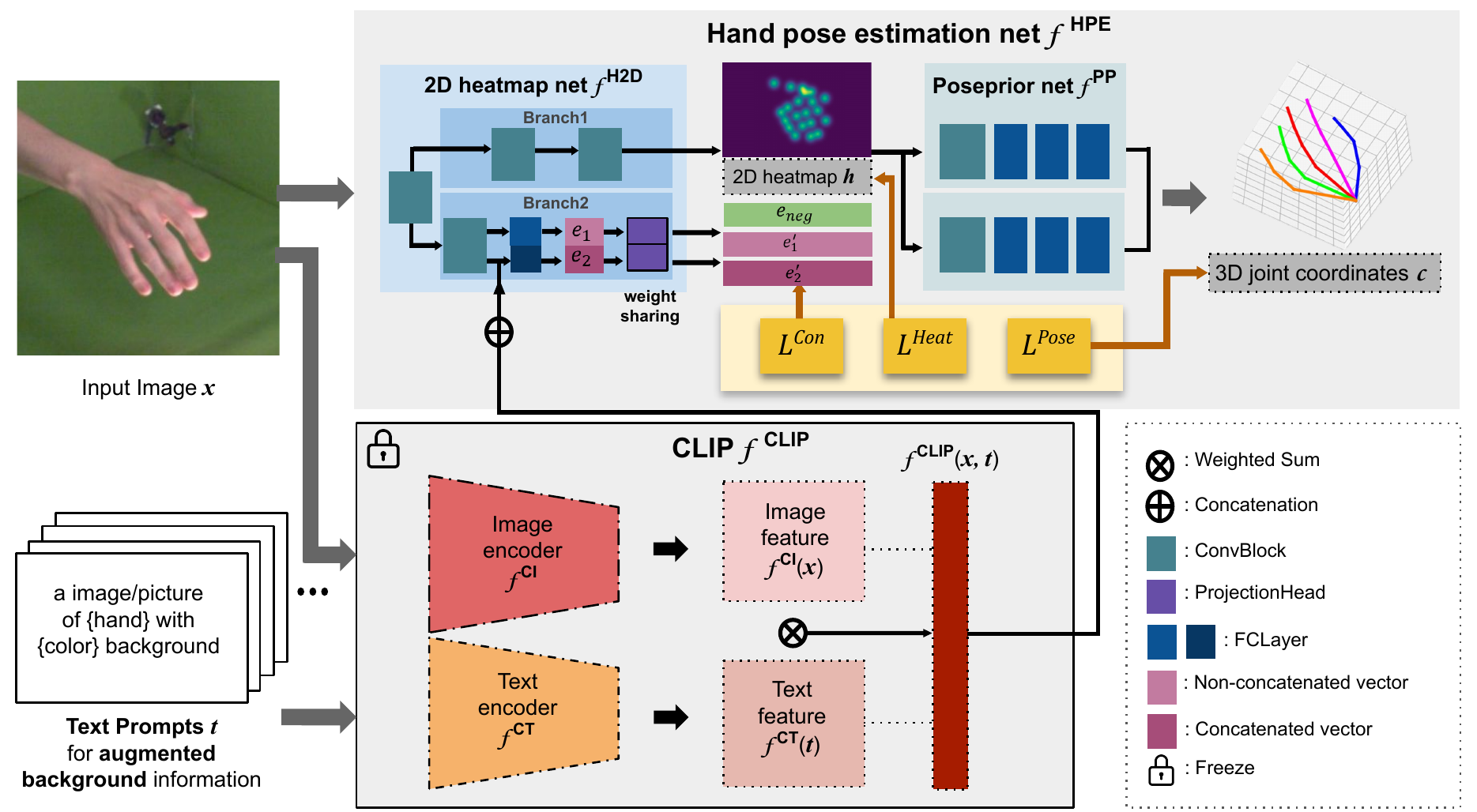}
\caption{A schematic diagram of of our domain generalization framework for 3D hand pose estimation. 
1)  The \textcolor{teal}{\textbf{CLIP network}} $f^{\text{CLIP}}$ receives the RGB image $\mbx$ and the text prompt $\mbt$ as inputs and generates a weighted sum of features $f^\text{CLIP}(\mbx, \mbt)$ from the CLIP image/text encoders, 2) \textcolor{violet}{\textbf{2D heatmap net}} $f^{\text{H2D}}$ receives the same RGB image $\mbx$ as input and generates the 2D heatmap $\mbh$ and encoding vectors $\mbe'_1$ and $\mbe'_2$. 3) \textcolor{purple}{\textbf{Poseprior net}} $f^{\text{PP}}$ receives the 2D heatmap $\mbh$ and generates 3D joint coordinates $\mbc$. In the 2D heatmap net $f^\text{H2D}$, the `Branch2' is taken to generate the encoding vector $\mbe'_1$ from the original sample and encoding vector $\mbe'_2$ combining the CLIP-augmented feature. Then, we applied the contrastive learning by regarding $\mbe'_1$ as the anchor, $\mbe'_2$ as the positive samples, and picking the negative sample $\mbe_\text{neg}$ based on the heatmap distance among samples in the same batch (See texts in Sec.~\ref{sec:train} for details). Via this, we are able to make our feature space of the 2D heatmap net $f^\text{H2D}$ robust to various domains. Three losses (ie. $L^\text{Con}$, $L^\text{Heat}$ and $L^\text{Pose}$) are used to train our hand pose estimation net $f^\text{HPE}$. 
} \label{fig:myfigure}
\end{figure*}


\section{Our hand domain generalization framework}
In this section, we introduce our hand pose estimation method utilizing CLIP model and contrastive learning mechanism for domain generalization. Overall, Our domain generalization framework receives an $256\times256\times3$-sized single RGB image $\mbx\in X$ as input and outputs $21\times 3$-dimensional 3D joint coordinates $\mbc\in C$. In the remainder of this section, we will first explain our baseline hand pose estimation network $f^\text{HPE}$, then explain about the CLIP network $f^\text{CLIP}$ and finally describe how we combine CLIP network $f^\text{CLIP}$ into our hand pose estimation network $f^\text{HPE}$ to construct the overall domain generalization framework. The overall schematic diagrams of our framework is shown in the Figure~\ref{fig:myfigure} and a list of used notations is provided in Table~\ref{t:notations}.

\begin{table}[]
\caption{Summary of notations.} 
\label{t:notations}
\resizebox{\linewidth}{!}{
\begin{tabular}{|l|l|}
\hline
$X \subset \mathbb{R}^{256\times 256\times 3}$ & RGB image space.\\ \hline
$T \subset \mathbb{R}^{3,920\times 1}$ & Text prompt space.\\ \hline
$E \subset \mathbb{R}^{128\times 1}$ & Encoding vector space.\\ \hline
$H \subset \mathbb{R}^{21\times 32\times 32}$ & 2D heatmap space. \\ \hline
$C \subset \mathbb{R}^{21\times 3}$ & 3D joint coordinate space. \\ \hline
$f^{\text{CLIP}}: [X, T] \to \mathbb{R}^{512\times 1}$ & CLIP model.  \\ \hline
$f^{\text{CI}}: X\to \mathbb{R}^{512\times 1}$ & CLIP image encoder. \\ \hline
$f^{\text{CT}}:T\to \mathbb{R}^{512\times 1}$ & CLIP text encoder. \\ \hline
$f^{\text{HPE}}: X\to C $ & Baseline 3D hand pose estimator. \\ \hline
$f^{\text{H2D}}: X\to [H, E] $ & 2D heatmap net. \\ \hline
$f^{\text{PP}}: H\to C$ & Poseprior net. \\ \hline
\end{tabular}
}
\end{table}

\subsection{Baseline 3D hand pose estimator $f^\text{HPE}$}
We constitute our baseline hand pose estimator $f^{\text{HPE}}$ samely as that of~\cite{zimmerman_iccv2017} which consists of 1) a 2D heatmap net $f^{\text{H2D}}$ that estimates the 2D heatmap $\mbh\in H$ from input RGB image $\mbx\in X$, and (2) a Poseprior net $f^{\text{PP}}$ that uses the estimated heatmap $\mbh\in H$ to predict 3D joint coordinates $\mbc\in C$. In the remainder of this subsection, we will explain details for two sub-networks (ie. $f^\text{H2D}$, $f^\text{PP}$).

\noindent \textbf{2D heatmap net $f^\text{H2D}$.} The 2D heatmap network $f^\text{H2D}$ receives a $256\times256\times3$-sized single RGB image $\mbx$. Then, it generates 21 $32\times32$-dimensional 2D heatmap $\mbh$. The input image $\mbx$ is sequentially applied to multiple ConvBlocks through the `Branch1' described in the Fig.~\ref{fig:myfigure} and it is mapped to the 2D heatmap $\mbh\in H$. The `Branch2' operations in Fig.~\ref{fig:myfigure} is related to involving CLIP features and this will be described in Sec.~\ref{sec:text_augm}. We use the convolutional pose machine (CPM) architecture~\cite{cpm_cvpr_2016} for constructing our 2D heatmap network $f^{\text{H2D}}$ following~\cite{zimmerman_iccv2017} and the detailed operations composing the architecture of the 2D heatmap network $f^\text{H2D}$ is described in the supplemental.

\noindent \textbf{Poseprior net $f^{\text{PP}}$.} After estimating the heatmap $\mbh\in H$ from the 2D heatmap net $f^{\text{H2D}}$, the Poseprior network $f^{\text{PP}}$ receives the estimated heatmap $\mbh$ and estimates corresponding 3D joint coordinates $\mbc\in C$. We train our model to predict the relative 3D coordinates within a given image frame and then converts them to absolute 3D coordinates as in~\cite{zimmerman_iccv2017}. In details, the network consists of two parallel processing streams. One stream predicts 3D hand pose in the canonical space. Another stream predicts a rotation matrix so that the 3D hand pose can be aligned in the camera space.  

\subsection{CLIP (Contrastive Language-Image Pre-Training) network $f^{\text{CLIP}}$}
The CLIP network $f^\text{CLIP}$ receives a 256$\times$256$\times$3-sized single RGB image $\mbx\in X$ and the text prompt $\mbt\in T$. The text encoder $f^\text{CT}$ and image encoder $f^\text{CI}$ of the CLIP model $f^\text{CLIP}$ generates a $512$-dimensional encoder features $f^{\text{CI}}(\mbx)\subset\mathbb{R}^{512\times 1}$ and $f^{\text{CT}}(\mbt)\subset\mathbb{R}^{512\times 1}$, respectively. In the original CLIP, ResNet-50 or ViT~(VisionTransformer) is used as the baseline structure for the image encoder; also Transformer based on the byte pair encoding is used for the text encoder. In this paper, we used the ViT-B/32 as the image encoder and the Transformer network as the text encoder, which are the pre-trained models provided by OpenAI \footnote{https://openai.com/} (We freeze the weights of CLIP model during the training).


\subsection{CLIP-augmented feature encoding}
\label{sec:text_augm}
Our CLIP-augmentation method is described in this section. We actually have two stages for this: 1) CLIP feature generation and 2) encoding vector generation stages.

\noindent \textbf{CLIP feature generation.}
The image $\mbx$ is input to the CLIP image encoder $f^\text{CI}$ to extract the feature vector $f^{\text{CI}}(\mbx)$. This vector contains rich information from the task-agnostic CLIP. However, it does not have information about images in other domains; while only having the information about images $\mbx$ that come from the source domain. To augment extra information from style and context, we proposed Table~\ref{tab:textprompt} to generate text prompt $\mbt$ reflecting diverse aspects of hand pose images. We can create text prompts by simply combining words in Table~\ref{tab:textprompt}. This method can generate $3,920$ text prompt configurations and some text prompts automatically generated are exampled in the supplemental. While more sophisticated text prompts could be made for each image and it would further improve the performance; we demonstrated that this simple method works well for our pipeline.

The generated text prompt $\mbt$ is used as the input to the CLIP text encoder $f^{\text{CT}}$ to extract the feature vector $f^{\text{CT}}(\mbt)$. To mix up the information, we defined $f^{\text{CLIP}}(\mbx, \mbt)$ as the weight-summed vector of $f^{\text{CI}}(\mbx)$ and $f^{\text{CT}}(\mbt)$. We chose the ratio between image encoder and text encoder as 6:4 and 9:1 for STB and RHD, respectively via the $10$-fold cross-validation. Ablation experiments on the ratio for weight-summation are shown in Table~\ref{t:ablation}(a). Via this process, we enforce the CLIP feature $f^{\text{CLIP}}(\mbx,\mbt)$ to contain the extra information in addition to the source domain information.

\begin{table}[!htb]
\caption{Compsition of text prompts}
\label{tab:textprompt}
\resizebox{1\columnwidth}{!}{%
\begin{tabular}{lcccccc}

\hline
& head               & hand color  & hand            & \multicolumn{2}{c}{color} & background \\ \hline
& a cropped image of & white       & hand with       & mountain     & lake       & room       \\ 
& a image of         & dark brown  & right hand with & bright       & dark       & background \\ 
& a cropped photo of & peach       &                & green        & purple     &            \\ 
& a picture of       & brown       &                & white        & yellow     &           \\ 
& one                & pale yellow &                & sky blue     & black      &           \\ 
& a photo of         & light beige &                & orange       & red        &           \\ 
& a photo of right   & black       &             & blue         & yellow     &           \\ 
&                &            &                & gray         & beige      &          \\ 
&                   &            &                & pink         & brown      &           \\ 
&                   &            &              & dotted       & flower     &           \\ 
\end{tabular}
}
\end{table}

\noindent \textbf{Encoding vector generation.} Given the CLIP feature $f^{\text{CLIP}}(\mbx,\mbt)$ extracted from the source image $\mbx$ and the text prompt $\mbt$, the 2D heatmap network $f^\text{H2D}$ receives a $256\times256\times3$-sized single RGB image $\mbx$ and generates $128$-dimensional encoding vectors $\mbe'_1\in E$ and $\mbe'_2\in E$ via the `Branch2' operations of 2D heatmap net $f^\text{H2D}$ (see Fig~\ref{fig:myfigure}). In this branch, the feature vector of the 2D heatmap network $f^\text{H2D}$ is sequentially applied to ConvBlock and FC layer to generate the $512$-dimensional  intermediate feature $\mbe_1$. Also, the feature vector of the 2D heatmap network $f^\text{H2D}$ is applied to ConvBlock and is concatenated with the CLIP feature $f^\text{CLIP}(\mbx, \mbt)$ and applied again to the FC layers to generate the $512$-dimensional intermediate feature $\mbe_2$. Intermediate features $\mbe_1$ and $\mbe_2$ are applied to the same (weight-shared) ProjectionHead layer to generate the $128$-dimensional encoding vectors $\mbe'_1$ and $\mbe'_2$, respectively. Here, the intermediate feature $\mbe_2$ is the enriched version of the intermediate feature $\mbe_1$ in the aspect of context and backgrounds by concatenating source domain information with the CLIP features $f^\text{CLIP}(\mbx, \mbt)$. The ProjectionHead is composed of two MLP layers that first projects $512$-dimensional vectors into $512$-dimensional vectors and then projects them again into the $128$-dimensional vectors.

Afterwards, the encoding vectors $\mbe'_1$ and $\mbe'_2$ are further exploited for the contrastive learning mechanism using the Eq.~\ref{eq:contrastiveloss}. The entire pipeline is trained in the end-to-end manner through 2D heatmap loss, 3D pose loss and contrastive loss. Via the contrastive loss, our pipeline becomes robust to images from various domains.

\subsection{Training}
\label{sec:train}

Our domain generalization framework is composed of end-to-end trainable networks based on 1) input image $\mbx$ from the source domain and 2) custom-created text prompt $\mbt$. We used only the source dataset (ie. FreiHAND~\cite{zimmermann2019FreiHAND}) for training and did not involve additional images or labels from the target domain or other unconstrained datasets. We trained the overall framework using three losses (ie. $L^{\text{Heat}}$, $L^{\text{Pose}}$ and $L^{\text{Con}}$) as follows:
\begin{eqnarray}
L(f^{\text{HPE}})&=&\lambda_1 L^{\text{Heat}} + \lambda_2 L^{\text{Pose}} + \lambda_3 L^{\text{Con}}
\end{eqnarray}
where $\lambda_1$, $\lambda_2$ and $\lambda_3$ are balance parameter controlling the weight of each loss function. From the $10$-random fold cross validation, we set 
$\lambda_1$ and $\lambda_2$ as $1$ and set $\lambda_3$ as $0.1$. Also, we used the Adam optimizer with $\beta=(0.9, 0.999)$ and a learning rate of $10^{-4}$. In the remainder of this section, we will explain about three losses.
 
\noindent \textbf{2D heatmap loss $L^{\text{Heat}}$.} The 2D heatmap loss  $L^\text{Heat}$ is defined as the standard mean square error (MSE) loss to close the predicted heatmaps $f^\text{H2D}(\mbx)=\mbh$ to their corresponding ground-truth 3D joint coordinates $\mbh^\text{GT}$ as follows:
\begin{eqnarray}
L^\text{Heat}(f^\text{H2D}) = \|f^\text{H2D}(\mbx)-\mbh^\text{GT}\|^2_2. 
\end{eqnarray}

\noindent \textbf{3D pose loss $L^{\text{Pose}}$.} The 3D pose loss  $L^\text{Pose}$ is also defined as the standard mean square error (MSE) loss as follows:
\begin{eqnarray}
L^\text{Pose}(f^\text{H2D}, f^\text{PP}) = \|f^\text{PP}(f^\text{H2D}(\mbx))-\mbc^\text{GT}\|^2_2
\end{eqnarray}
where $\mbc^\text{GT}$ denotes the ground-truth 3D joint coordinates.

\noindent \textbf{Contrastive loss $L^{\text{Con}}$. } 
The contrastive loss is employed to maximize the latent space agreement with the positive samples while minimizing the agreement with negative samples. The encoding vector $\mbe'_1$ becomes the anchor while the encoding vector $\mbe'_2$ becomes the positive sample. After this, among the samples in the same batch, the negative sample $\mbe_\text{neg}$ is selected as the sample whose ground-truth heatmap is farthest from the predicted heatmap of the anchor sample. The contrastive loss $L^\text{Con}$ is then defined as follows:
\begin{eqnarray}
\label{eq:contrastiveloss}
L^\text{Con}(f^\text{H2D}) = \|\mbe'_1-\mbe'_2\|_1 - \max(0.5, \|\mbe'_1-\mbe_\text{neg}\|_1).
\end{eqnarray}

\subsection{Testing}

Only the hand pose estimation net $f^{\text{HPE}}$ disregarding the `Branch2' of 2D heatmap net $f^\text{H2D}$ is exploited during the testing stage. The test RGB image $\mbx$ is input to the 2D heatmap net $f^\text{H2D}$ and it takes the `Branch1' route (see Fig.~\ref{fig:myfigure}) to generate 21 $32\times 32$-dimensional heatmap $\mbh$. Then the poseprior net $f^\text{PP}$ is further applied to map it towards the $21\times 3$-dimensional 3D joint coordinates $\mbc$. 


\section{Experiment}

In this section, we elaborate our experimental settings and analyze the results qualitatively and quantitatively. We demonstrate that the domain generalization method utilizing CLIP has better generalization capability than the previous state-of-the-art methods.

\subsection{Dataset}

We conduct our experiments using three types of RGB-based 3D hand pose benchmark datasets which have RGB images and corresponding ground-truth 3D pose annotations.

\noindent \textbf{FreiHAND dataset.} FreiHAND \cite{zimmermann2019FreiHAND} is a large 3D hand dataset consisting of 130,240 training images and 3,960 testing images. This dataset also includes the mesh annotation as well as the hand pose annotation. Testing data are created both having indoor and outdoor environments; while training data consist of data taken in an indoor environment with a green background. Afterwards, it provides training data which are artificially synthesized with the background. We use this dataset as the source domain data.

\noindent \textbf{STB dataset.} Stereo Hand Pose Tracking Benchmark(STB)~\cite{mueller2018ganerated} provides 2D and 3D annotations of 21 keypoints with a resolution of $640\times 480$. It is a real dataset, composed of STB-BB, STB-SK subsets of images. Two different subsets are captured by the Point Grey Bumblebee2 stereo camera and the Intel F200 depth camera, respectively. We follow the training and testing splits of Zimmerman~\etal~\cite{zimmerman_iccv2017}~($15,000$ images for training and $3,000$ images for testing) and use only the test splits as the target domain dataset in our experiments.

\noindent \textbf{RHD dataset.}
Rendered hand pose dataset(RHD)~\cite{zimmerman_iccv2017} is a synthetic dataset captured by \emph{Blender} software using $20$ different characters from Mixamo performing $39$ actions. It has overall $43,986$ images with a resolution of $320\times 320$ pixels and accurate $21$ keypoints annotations and segmentation masks. The background images having cities and landscapes are randomly sampled from \emph{Flickr}. We follow the training and testing splits of Zimmerman~\etal~\cite{zimmerman_iccv2017} ($41,258$ images for training, $2,728$ images for testing) and use only the test splits as target domain dataset in our experiments.

\noindent \textbf{Evaluation method.}
We evaluated the proposed algorithm on two hand pose estimation datasets (ie. STB and RHD) based on the end-point-error (EPE) measure that calculates the distance between estimated 3D joint coordinates $\mbc$ and ground-truth 3D joint coordinates $\mbc^\text{GT}$ in the $mm$ unit.

\subsection{Results.}

This paper deals with the domain generalization (DG) problem in the 3D hand pose estimation task. We mainly compared our method with the existing DG method: Zhang~\etal~\cite{zhang2021learning} that solves the same problem with ours and achieves the best performance before us. We involved their results~\cite{zhang2021learning} for involving only `source domain images'. In Table~\ref{t:comparison}, we compared ours with existing methods and confirm that our method showed the superior performance based only on the source images, without involving target images (ie. STB, RHD datasets) or additional unconstrained images. Compared to Zhang~\etal~\cite{zhang2021learning}, via our method, error rates are decreased by 3.33\%  and 3.11\% in STB and RHD datasets, repectively.

\begin{figure}[ht]
\centering
\includegraphics[width=\linewidth]{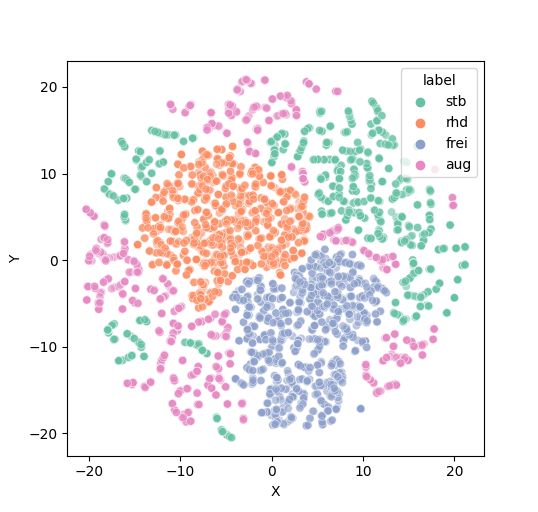}
\caption{Visualization of T-SNE distribution obtained using CLIP image and text encoder features from source dataset (FreiHAND) (ie. `aug') and CLIP image features from existing datasets (ie. `stb' for STB, `rhd' for RHD and `frei' for FreiHAND datasets).} \label{fig:myfigure2}
\end{figure}

Figure~\ref{fig:myfigure2} shows the visualization of T-SNE distribution for the CLIP image and text encoder features (ie. $0.5\times (f^\text{CI}(\mbx)+f^\text{CT}(\mbt))$) obtained from samples in the source dataset (FreiHAND) and the CLIP image encoder feature (ie. $f^\text{CI}(\mbx)$) obtained from samples in the existing datasets (ie. STB, RHD and FreiHAND). The `aug' denotes the samples obtained from the CLIP image and text encoder features (ie. $0.5\times (f^\text{CI}(\mbx)+f^\text{CT}(\mbt))$) for the source dataset (FreiHAND); while `stb', `rhd' and `frei' denote samples obtained from the CLIP image encoder feature (ie. $f^\text{CI}(\mbx)$) for STB, RHD and FreiHAND datasets, respectively. Figure~\ref{fig:myfigure2} shows that the distribution of augmented features (ie. `aug' samples) cover the distribution of STB and RHD datasets, which are target domain datasets. Through this experiment, we qualitatively visualize the effect of our proposed method.

\begin{figure*}[ht]
\centering
\includegraphics[width=\linewidth]{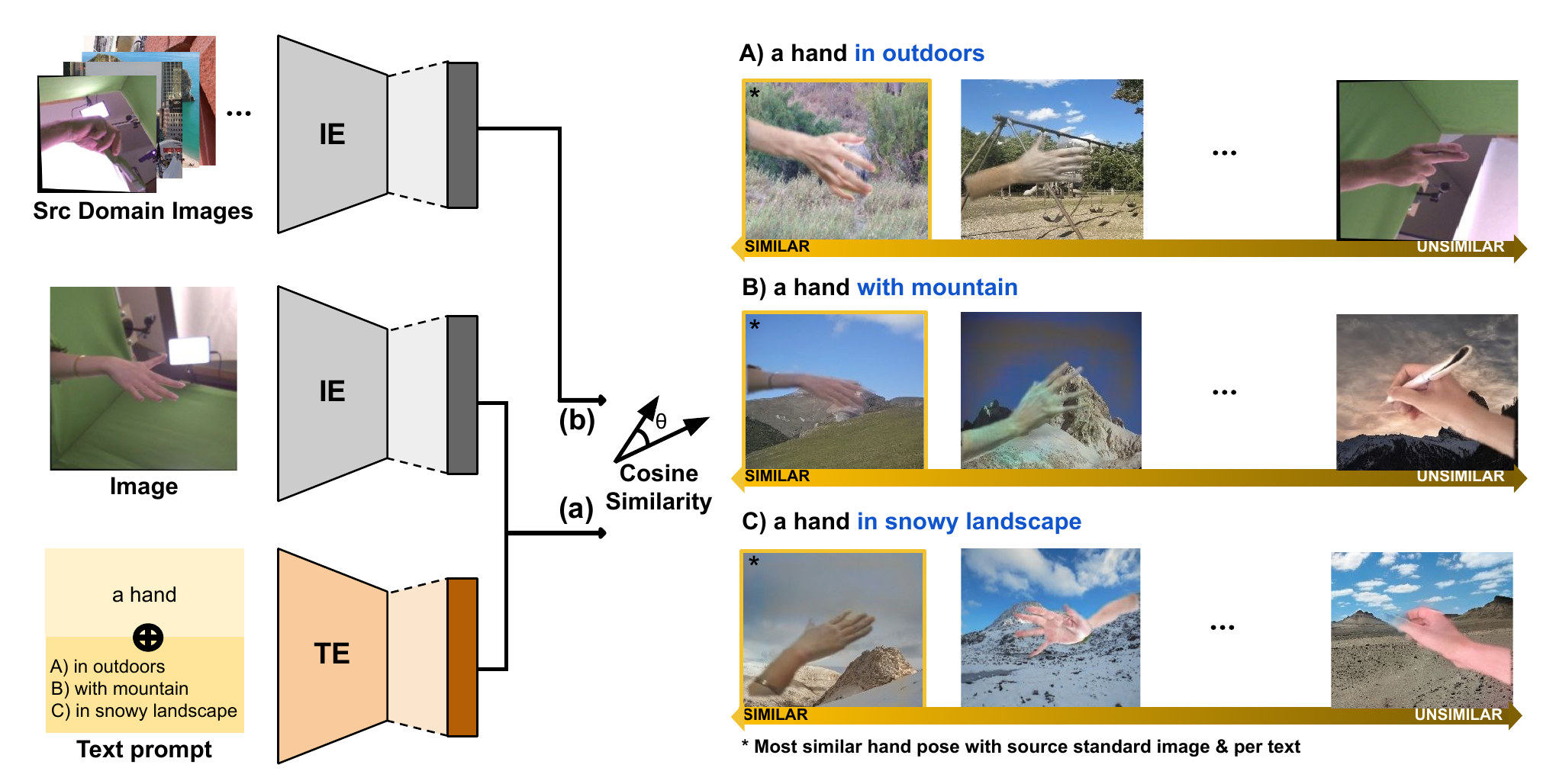}
\caption{A visualization of FreiHAND images which has similar features to CLIP-augmented features. When the image of the source domain (FreiHAND) and the text prompt for the augmentation are given (left), in (a), we first weight-sum the features obtained from CLIP image encoder (IE) and CLIP text encoder (TE). After that, for (b), we apply the CLIP image encoder to all FreiHAND source data without the input image used in (a). After calculating the cosine similarity between features obtained in (a) and (b), images could be ranked according to their similarity and ranked images are visualized in the right side of this figure. The ranked images are seemingly well aligned with the text prompts..}
\label{fig:myfigure3}
\end{figure*}

The validity of the augmented distribution is proven in
Figure~\ref{fig:myfigure3} which shows that our method maintains the context information of source domain and style of text prompt. In Figure~\ref{fig:myfigure3}, the image of the source domain dataset (FreiHAND) and text associated with the image are given to the CLIP image encoder (IE) $f^\text{CI}$ and CLIP text encoder (TE) $f^\text{CT}$, respectively and transformed to output features. And then, the two output features (ie. image feature and text feature) are merged  by the weight summation (a). The rest of the images in the source domain dataset, except for the input image are given to the image encoder to extract image features (b). After extracting features, the cosine similarities between (a) and (b) are calculated. In the right side of the Fig.~\ref{fig:myfigure3}, we visualized text prompts with source domain images ranked by their similarity. The samples having the highest cosine similarity might be most relevant samples to the combined input image $\mbx$ and text prompt $\mbt$. From the visualization, we could observe that such an assumption is valid.

\begin{table}[h!]
\centering
\resizebox{\columnwidth}{!}{%
\begin{tabular}{c|c|c}
\hline
\multirow{2}{*}{Methods}              & FreiHAND→STB     & FreiHAND→RHD \\ \cline{2-3} 
                                      & EPE↓             & EPE↓         \\ \hline
Zhang~\etal~\cite{zhang2021learning}  & 36.1             & 48.3       \\ \hline
Ours                                  & \textbf{34.9}   & \textbf{46.8}         \\ \hline
\end{tabular}%
}
\caption{Comparison with the state-of-the-art methods on STB, RHD datasets. Units are in mm scale.}
\label{t:comparison}
\end{table}

\begin{table}[h!]
\begin{center}
\begin{tabular}{ cc }
\begin{tabular}{cc|c|c}
\hline
\multicolumn{1}{c|}{\multirow{2}{*}{Methods}}& \multirow{2}{*}{Ratio} & STB   & RHD            \\ \cline{3-4} 
\multicolumn{1}{c|}{}                         &                        & EPE↓  & EPE↓           \\ \hline
\multicolumn{2}{c|}{Baseline}                                          & 36.11 & 48.36          \\ \hline
\multicolumn{2}{c|}{IE}                                                & 35.00 & 48.37          \\ \hline
\multicolumn{1}{c|}{\multirow{5}{*}{IE+TE}}   & 0.9                    & 36.69 & \textbf{46.82}          \\ \cline{2-4} 
\multicolumn{1}{c|}{}                         & 0.8                    & 35.75 & 49.43 \\ \cline{2-4} 
\multicolumn{1}{c|}{}                         & 0.6                    & \textbf{34.97} & 47.34          \\ \cline{2-4} 
\multicolumn{1}{c|}{}                         & 0.4                    & 47.25 & 63.01          \\ \cline{2-4}
\hline
\end{tabular}
\\
(a)
\\
\\
\begin{tabular}{c|c|c|c}
\hline
\multirow{2}{*}{Methods} &  Combination              & STB  & RHD  \\ \cline{2-4}  
                         &     -           & EPE↓ & EPE↓ \\ \hline
{Baseline}                 &      -          & 36.11 & 48.36 \\ \hline
\multirow{3}{*}{TE}      & Concatenation           &38.12      & 49.46      \\ \cline{2-4} 
                         & Summation      & 35.93 & 48.36 \\ \cline{2-4} 
                        \hline
\multirow{3}{*}{IE+TE}   & Concatenation           & \textbf{34.97} & \textbf{47.34} \\ \cline{2-4}  
                         & Summation      & 42.98 & 48.24 \\ \cline{2-4} 
                        \hline 
\end{tabular} &
\end{tabular} \\
(b)
\end{center}
\caption{Ablation study for hyper-parameters. (a) We compared results for various ratios of CLIP encoders. Here, the ratio indicates the weight assigned to the CLIP image encoder $f^\text{CI}$, 1-ratio is the weight assigned to the text encoder $f^\text{CT}$. The weight summation using the ratio of $0.6$ and $0.9$ works best for STB and RHD datasets, respectively. (b) We compared the combination methods among the concatenation and summation operations. We could find that the concatenation method works better than summation method.}
\label{t:ablation}
\end{table}

\begin{table}[]
\begin{center}
\begin{tabular}{c|c|c|c}
\hline
\multirow{2}{*}{Methods} & \multirow{2}{*}{$\lambda_3$} & STB  & RHD  \\ \cline{3-4} 
                        &                                   & EPE↓ & EPE↓ \\ \hline
\multirow{4}{*}{IE+TE}  & 1                               &      37.67 &   53.14   \\ \cline{2-4} 
                        & 0.1                             &  34.97 & \textbf{47.34}   \\ \cline{2-4} 
                        & 0.01                             &      35.05 & 50.92     \\ \cline{2-4} 
                        & 0.001                            &      \textbf{30.65} & 50.97     \\ \hline
\end{tabular}
\end{center}
\caption{(a) A comparison of our methods under various $\lambda_3$ values of the contrastive loss. The best $\lambda_3$ value is set as $0.001$ and $0.1$ for STB and RHD datasets, respectively. However from the cross-validation, we set our $\lambda_3$ as $0.1$ and report it as the final accuracy. }
\label{t:ablation_ratio}

\end{table}

\begin{table}[]
\begin{center}
\begin{tabular}{c|c|c}
\hline
\multirow{2}{*}{Methods}                                       & STB           & RHD           \\ \cline{2-3} 
                                                              & EPE↓          & EPE↓          \\ \hline
Ours                                                          & \textbf{34.97} & \textbf{46.82} \\ \hline
\begin{tabular}[c]{@{}c@{}}Normal\\ Augmentation\end{tabular} & 38.92         & 49.86         \\ \hline
\end{tabular}
\end{center}
\caption{(a) A comparison of our method versus the normal augmentation method. At the normal augmentation method, there are six types of augmentation applied (ie. color jitter, cut out, gaussian noise, sobel filter, color drop, and gaussian blur). Ours consistently and significantly works better than the normal augmentation method. }
\label{t:normal_aug_compar}
\end{table}

\subsection{Ablation study.} We assess the contribution of the CLIP model applied to hand pose estimation in our method: We configure four variants in our framework:
1) We prove the effectiveness of the CLIP model on the domain generalization problem through comparisons with and without the CLIP network, and study the optimal ratio between $f^{\text{CI}}(\mbx)$ and $f^{\text{CT}}(\mbt)$ when generating the combined features $f^{\text{CLIP}}(\mbx,\mbt)$ from them in Table~\ref{t:ablation}(a). 2) For the optimal combination of hand pose estimator features and CLIP features, we study which operation is effective to combine CLIP features to the original features between `concatenation' and `summation'. We also obtained the results for the case when we only used the CLIP text encoder for applying the style augmentation in Table~\ref{t:ablation}(b). 3) We experimented to find the optimal weight ratio (ie. $\lambda_3$) between contrastive loss and other loss functions in Table~\ref{t:ablation_ratio}. 4) Finally, we compared the performance of our method and the method applied with normal data augmentation in Table~\ref{t:normal_aug_compar}. 
In Table~\ref{t:ablation}(a), we compared the performance of several variants: baseline, our method using only CLIP image encoder (IE) and our method using CLIP image encoder + CLIP text encoder (IE+TE). The error rate is reduced by $3.08\%$ in STB even when only CLIP image encoder is further involved. However, there is an significant improvement in both STB and RHD when both CLIP image encoder and CLIP text encoder are involved. In Table~\ref{t:ablation}(a), the optimal weight ratio between $f^{\text{CI}}(\textbf{x})$ and $f^{\text{CT}}(\textbf{t})$ is seemingly $0.6$ and $0.9$ in STB and RHD, respectively. We obtained the values from the $10$-fold cross-validation. In Table~\ref{t:ablation}(b), we study the effectiveness of IE+TE method on the style augmentation. From our results, we can observe that the sole CLIP text encoder (TE) is not enough to achieve the domain generalization to other domains since it does not have the context information from the source domain. On the other hand, `IE+TE' can maintain context information of source domain and augmented style information via both CLIP image encoder and CLIP text encoder, thereby achieving the complete domain generalization. We also study the effect of two operations when combining hand pose estimator features and CLIP features: concatenation and summation. Also, the results showed that the concatenation is more effective operation than the summation to incorporate CLIP features into the hand pose estimation features. 
In Table~\ref{t:ablation_ratio}, we showed results for the optimal weight between the contrastive loss and other loss functions (ie. $\lambda_3$). The best performance is obtained when $\lambda_3=0.001$ and $\lambda_3=0.1$ for STB and RHD, respectively in the table; while we set $\lambda_3=0.1$ for both datasets as the value is obtained from $10$-fold cross-validation. Thus, we report $\lambda_3=0.1$ results both for STB and RHD datasets. Finally, we compare the normal data augmentation method with ours in Table~\ref{t:normal_aug_compar}. We showed that our augmentation method consistently and significantly performs better than the normal data augmentation method. For the normal data augmentation, we involved six types of augmentation: color jitter, cut out, gaussian noise, sobel filter, color drop, and gaussian blur.

\section{Conclusion}
In this paper, we propose the 3D hand pose estimation framework that generalizes well to unseen domain datasets. Existing domain adaptation/generalization approaches tried to relieve the dataset-bias or domain-shift problem among different datasets by supplying additional unconstrained/target domain dataset during the training stage. However, it takes a lot of time and effort to collect such datasets or sometimes impossible to get the exact target domain distribution. In this paper, we proposed image-free domain generalization framework that involves the CLIP model for generating style-augmented feature by the text prompt which is related to hand domains.

The hand pose estimator becomes to have the style-augmented features through the contrastive learning mechanism, so the hand pose estimation network becomes robust to unseen domain data. In experiments, we trained our network on FreiHAND dataset and test on two popular hand pose estimation datasets (ie. RHD and STB). For two datasets, we have achieved the state-of-the-art performance in domain generalization setting by decreasing the error rates by $3.33\%$ and $3.11\%$ compared to previous state-of-the-art method in STB and RHD datasets, respectively. We demonstrated that the text prompt can augment the style information and makes our model to generalize to other domain datasets. We expect that our method can be extended to other tasks or in other pose estimation tasks (ie. human body pose and animal pose estimation).

\noindent \textbf{Acknowledgements.} This work was supported by IITP grants (No. 2021-0-01778 Development of human image synthesis and discrimination technology below the perceptual threshold 20\%; No. 2020-0-01336 Artificial intelligence graduate school program (UNIST) 20\%; No. 2021-0-02068 Artificial intelligence innovation hub 20\%; No. 2022-0-00264 Comprehensive video understanding and generation with knowledge-based deep logic neural network 20\%) and the NRF grant (No. 2022R1F1A1074828 20\%), all funded by the Korean government (MSIT).

{\small
\bibliographystyle{ieee_fullname}
\bibliography{egbib}
}

\end{document}